\documentclass[sn-mathphys,Numbered]{sn-jnl}% Math and Physical Sciences Reference Style
\usepackage{graphicx}%
\usepackage{multirow}%
\usepackage{amsmath,amssymb,amsfonts}%
\usepackage{amsthm}%
\usepackage{mathrsfs}%
\usepackage[title]{appendix}%
\usepackage{xcolor}%
\usepackage{textcomp}%
\usepackage{manyfoot}%
\usepackage{booktabs}%
\usepackage{algorithm}%
\usepackage{algorithmicx}%
\usepackage{algpseudocode}%
\usepackage{listings}%
\usepackage{rotating}%
\theoremstyle{thmstyleone}%
\theoremstyle{thmstyletwo}%

\theoremstyle{thmstylethree}%

\begin{document}

\title[Article Title]{MonoPIC - A Monocular Low-Latency Pedestrian Intention Classification Framework for IoT Edges Using ID3 Modelled Decision Trees}


\author[1]{\fnm{Sriram} \sur{Radhakrishna}}\email{sriram.radhakrishna42@gmail.com}

\author*[1]{\fnm{Adithya} \sur{Balasubramanyam}}\email{adithyab@pes.edu}

\author[1]{\fnm{Prasad} \sur{B Honnavalli}}\email{prasadhb@pes.edu}

\affil[1]{\orgdiv{Department of Computer Science and Engineering}, \orgname{PES University}, \orgaddress{\street{100 Feet Ring Road, Banashankari Stage III}, \city{Bengaluru}, \postcode{560086}, \state{Karnataka}, \country{India}}}


\abstract{%% Text of abstract
Road accidents involving autonomous vehicles commonly occur in situations where a (pedestrian) obstacle presents itself in the path of the moving vehicle at very sudden time intervals, leaving the robot even lesser time to react to the change in scene. Therefore, it is important to take certain factors nto account in pedestrian avoidance systems that influence the reation time by improving their latency.

In order to tackle this issue, we propose a novel algorithmic implementation that classifies the intent of a single arbitrarily chosen pedestrian in a two dimensional frame into logic states based on orientation changes represented in quaternion format. This bypasses the need to employ any relatively high latency deep-learning algorithms primarily due to the lack of necessity for depth perception as well as an implicit cap on the computational resources that most IoT edge devices present.

The model was able to achieve an average testing accuracy of 83.56\% with a reliable variance of 0.0042 while operating with an average latency of 48 milliseconds. This demonstrates multiple notable advantages over the current standard of using spatio-temporal convolutional networks for these perceptive tasks.}




\maketitle

\section{Introduction}

Pedestrian safety is a critical concern in the development and implementation of intelligent transportation systems (ITS). Even without the advent of autonomous vehicle technologies, pedestrian deaths have surged 18\% \cite{snider_2023} between the first half of 2019 and 2022. To enhance pedestrian safety, it is crucial to develop accurate methods to determine pedestrian intent with a quick response time. In this research paper, we present a decision tree-based approach for determining pedestrian intent. The proposed decision tree utilizes a combination of pedestrian features, such as motion patterns and behavioral cues, to determine pedestrian intent. The novel approach is designed to be robust, accurate, and computationally efficient, making it suitable for real-time implementation in ITS due to their impact on latency. To validate the approach, experiments were conducted using real-world pedestrian datasets and compared the results with existing methods \cite{mersch2021maneuver}\cite{wang2021human}\cite{pellegrini2009you}. The findings demonstrate the effectiveness of the approach in accurately determining pedestrian intent.

The proposed decision tree-based approach builds upon previous research in pedestrian intent prediction \cite{xin2022prediction} and offers several advantages over existing methods. Firstly, the approach is designed to incorporate a smaller but more impactful range of pedestrian features to improve the accuracy of pedestrian intent prediction. Secondly, the approach is computationally efficient, allowing for real-time implementation in ITS. This makes it particularly useful in high-density urban environments where pedestrian traffic can be unpredictable and requires rapid decision-making.

\begin{figure}[htbp]
\centerline{\includegraphics[scale=0.35]{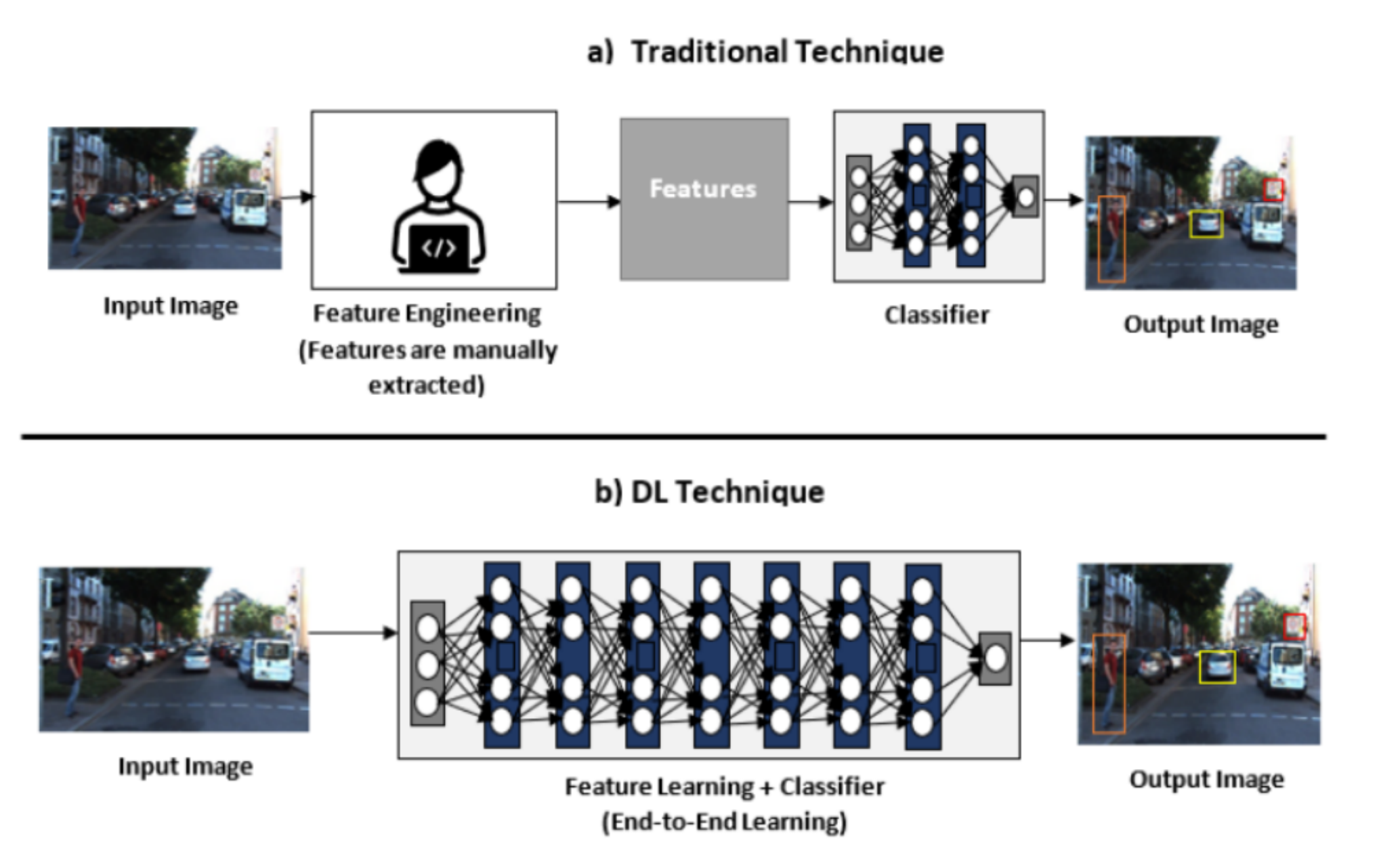}}
\caption{A representation of a deep learning model, commonly used to evaluate pedestrian intent. Sub-figure a involves the traditional method with manual feature engineering and classification \cite{turner1999conceptual} while sub- figure b illustrates a generic deep learning model given the task of both feature extraction and classification \cite{lecun2015deep}.}
\end{figure}

The experimental results demonstrate that the approach achieves high accuracy in predicting pedestrian intent in various scenarios. The approach was compared with existing methods, including temporal convolutional networks \cite{saleh2019real}  similar to the one referenced in figure 1.a and body pose analysis, and found that the approach outperformed these methods in terms of accuracy and computational efficiency. These findings indicate that the approach has significant potential to enhance pedestrian safety and reduce the risk of accidents involving pedestrians in ITS.

Overall, the proposed decision tree-based approach provides an effective and quick method for determining pedestrian intent. The results of the experiments demonstrate the effectiveness of the approach and its potential to improve pedestrian safety in ITS. The findings contribute to the growing body of research on pedestrian intent prediction and offer insights into the development of intelligent transportation systems that can effectively address pedestrian safety concerns.

To address the problem of latency, it is logical to minimize the computational workload and the number of calculations required by the algorithm in use. To achieve this objective, it is crucial to identify the specific inputs that are being processed by these models for assessing the danger presented by a pedestrian object\cite{gandhi2006pedestrian}. These inputs include the detection of the pedestrian, their orientation angle concerning the self-driving vehicle, the relative displacement, and the apparent velocity with other sensor data \cite{gandhi2006pedestrian}. These inputs essentially represent the spatio-temporal relationships that can be extracted from the image feed through convolutions. By removing intermediate components of the machine learning model, the computational effort can be reduced. 

\begin{figure}[htbp]
\centerline{\includegraphics[scale=0.5]{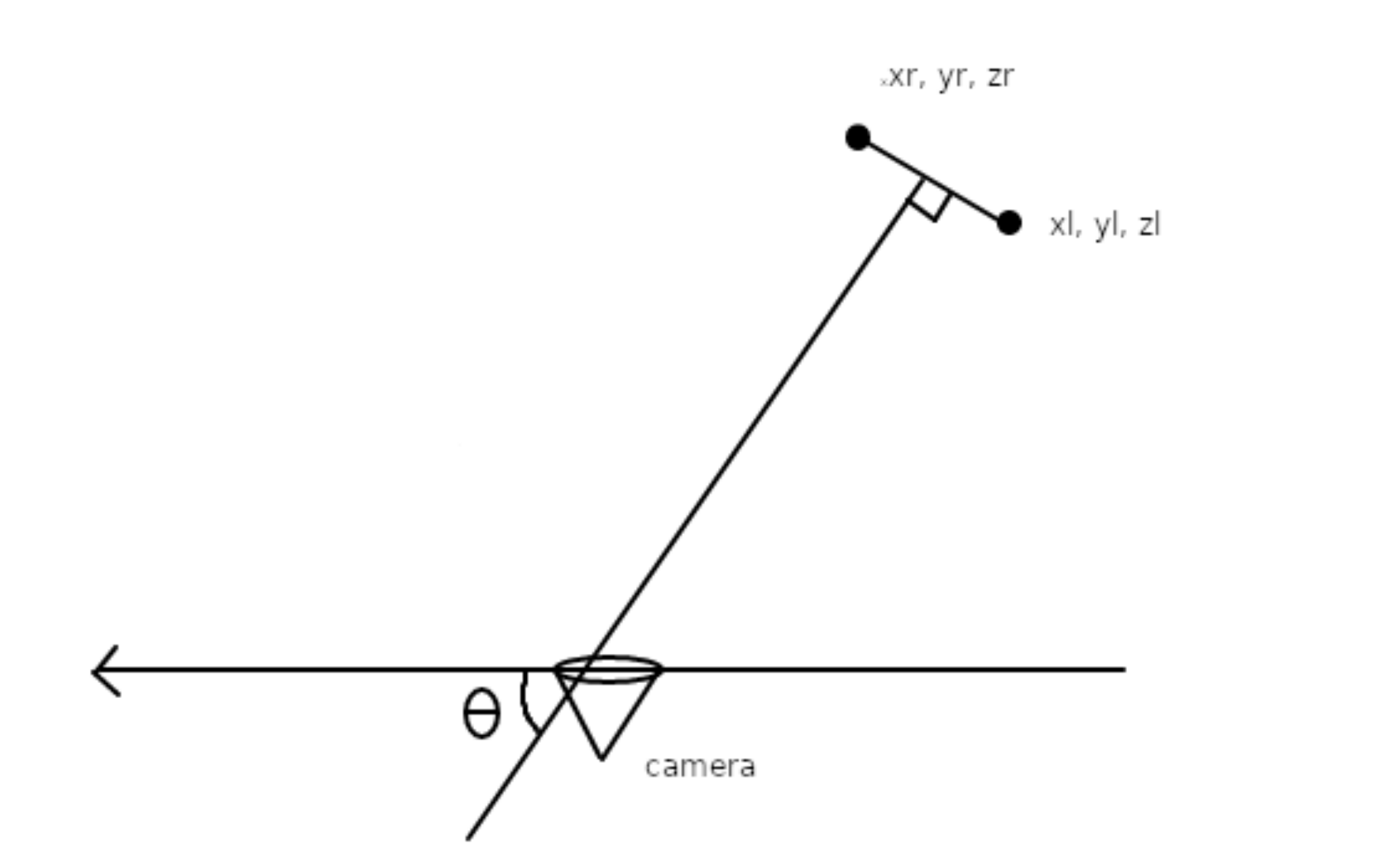}}
\caption{A simplified representation of the angle of orientation with the horizon being captured as referenced from \cite{radhakrishna2023economical}.}
\end{figure}

The difficulties encountered in this approach relate to applying algorithmic solutions to identify the implied inputs discussed earlier. The initial challenge involved determining the orientation of the pedestrian obstacle while maintaining acceptable latency levels for localizing body landmarks. To resolve this, a mathematical function was developed to extract the angle of orientation from the pose data of these landmarks. Secondly, the velocity of the pedestrian obstacle was established by extrapolating displacements along the x and y components of the image feed, measured in pixels. This approach restricted the frame of reference to the image, negating the requirement for depth data, similar in goal to recent works by Cao et al. \cite{cao2017realtime}.

In order to prove the efficacy of MonoPIC on the relevant metrics, a quantitative evaluation was conducted involving the following tests :

\begin{enumerate}
    \item Calculations per prediction - This test is crucial to evaluate the algorithm's efficiency. It helps in understanding how many computations MonoPIC performs to make a single prediction about pedestrian movement intent.
    \item Power consumption - This test aims to measure the power consumption of the device when MonoPIC is actively processing data. It's essential to ensure that the algorithm is energy-efficient for deployment on resource-constrained IoT edge devices.
    \item Latency comparison with other algorithms on single core - This test compares MonoPIC's response time (latency) with other algorithms designed for the same task. It provides insights into the algorithm's speed and responsiveness.
\end{enumerate}

The comprehensive testing of MonoPIC yielded insightful conclusions regarding its viability for deployment on IoT edge devices. The calculations per prediction analysis highlighted the algorithm's computational efficiency, indicating its ability to swiftly process information without excessive computational overhead. The power consumption tests in a single-threaded context, demonstrated MonoPIC's capacity to maintain energy efficiency, a critical factor for edge devices with limited resources. Moreover, the latency comparison against equivalent algorithms underscored MonoPIC's competitive speed and responsiveness in discerning pedestrian movement intent. These findings collectively suggest that MonoPIC is not only adept at making accurate predictions but is also optimized for resource utilization, making it a promising candidate for real-world applications on IoT edge devices. The agnostic nature of these tests, devoid of dependence on specific datasets, enhances the generalizability of these conclusions across diverse scenarios, affirming MonoPIC's robust performance across varied operational conditions.

This paper will cover the theoretical aspects of the implementation of MonoPIC with the reasoning behind its overarching design decisions, followed by an analysis of the advantages it brings to the table.

\section{Related work}

To enhance the transparency of the evaluation process, it's crucial to outline the methods employed for literature review and testing.
The literature review encompassed studies and publications spanning the last five years. This timeframe ensured the incorporation of recent advancements in algorithms for estimating pedestrian movement intent, aligning with the dynamic nature of the field.
The search strategy involved a comprehensive selection of keywords to capture the breadth of relevant literature. Key terms included "pedestrian movement intent," "edge computing," "IoT devices," "algorithm efficiency," and "power consumption optimization." This diverse set of keywords aimed to retrieve studies that covered algorithmic aspects, deployment on edge devices, and energy efficiency considerations.
The literature review focused on addressing specific research questions to inform the testing parameters:
\begin{enumerate}
    \item Efficiency: What are the latest advancements in algorithms for estimating pedestrian movement intent, and how computationally efficient are they?
    \item Edge Deployment: How do these algorithms fare when deployed on IoT edge devices in terms of power consumption?
    \item Performance Comparison: How does the latency of MonoPIC compare with equivalent algorithms designed for the same task?
\end{enumerate}

By systematically addressing these research questions through the literature review, the testing framework was designed to assess MonoPIC's capabilities in alignment with the state-of-the-art developments in the field. This methodological approach ensures that the evaluation is grounded in the current landscape of pedestrian movement intent estimation algorithms and their applicability to edge computing scenarios.

Pedestrian intent detection is an essential aspect of road safety, and several studies have explored different approaches to address this challenge. One study proposed a multi-scale pedestrian intent prediction method using 3D joint information as spatio-temporal representation, which does not require a detector and improves the run-time of previous approaches \cite{ahmed2023multi}. Another study presented a skeleton-based prediction approach for pedestrian intent detection using a camera feed and the Mediapipe framework to isolate the skeletal structure of the detected human \cite{samant2021pedestrian}.

The future of pedestrian intent prediction is also discussed in a report, which highlights the importance of pedestrian intent in autonomous vehicles (AVs) to navigate safely and avoid fatalities/injuries \cite{Netscribes_2023}. The report discusses the use of AI vision-based systems and deep neural networks to infer pedestrian intention, with a focus on annotated learning to execute high-performing actions.

Additionally, a research paper on pedestrian intent prediction using deep machine learning highlights the high complexity of pedestrian intent prediction due to the agility of pedestrians in their movement and the potential of deep machine learning models in addressing this challenge \cite{bapu2020pedestrian}. 

These studies demonstrate the various approaches and techniques employed in pedestrian intent detection, including multi-scale prediction, skeleton-based prediction, AI vision-based systems, and deep machine learning models. Further research is required to develop more robust and accurate models for pedestrian intent detection, with a focus on real-time implementation and improving pedestrian safety.

Decision trees are a popular technique for pedestrian intent prediction in intelligent transport systems. This study proposes a novel algorithmic implementation that classifies the intent of a single arbitrarily chosen pedestrian in a two-dimensional space using ID3 modelled decision trees for IoT edge devices. Another study proposed a pedestrian crossing intention prediction method based on multi-feature fusion using a random forest model \cite{ma2022pedestrian}. A comprehensive survey on pedestrian intention prediction for autonomous vehicles also highlights the importance of decision trees in this field\cite{sharma2022pedestrian}. Additionally, a study investigated the prediction of pedestrians' wait-or-go decision using trajectory data based on gradient boosting decision tree \cite{xin2022prediction}. These studies demonstrate the effectiveness of decision tree-based algorithms in predicting pedestrian intent using various features and models.

The prediction of pedestrian intent is crucial for safe and efficient operation of intelligent transport systems. One of the popular techniques for pedestrian intent prediction is the usage of decision trees. Decision trees can model complex decision-making processes by recursively partitioning the input space based on a set of rules. Several studies have utilized decision trees for pedestrian intent prediction in various settings for robotic decision making.

Another study presented a pedestrian crossing intention prediction method based on multi-feature fusion using a random forest model \cite{ahmed2023multi}. These studies demonstrate the effectiveness of decision tree-based algorithms in predicting pedestrian intent using various features and models.

The use of decision trees for pedestrian intent prediction showcases their potential in addressing the challenges of intelligent transport systems. Further research in this area is essential to develop more robust and accurate decision tree-based models, with a focus on improving latency and real-time prediction capabilities.

Additionally, a research paper on pedestrian intent prediction using deep machine learning highlights the high complexity of pedestrian intent prediction due to the agility of pedestrians in their movement and the potential of deep machine learning models in addressing this challenge \cite{bapu2020pedestrian}.

MediaPipe is an open-source framework maintained by Google, offering a comprehensive set of tools, APIs, and pre-trained models that make building applications for tasks like pose estimation, object detection, facial recognition, and more, easier than ever. One of the key tasks provided by MediaPipe is the Pose Landmarker, which allows the detection of landmarks of human bodies in an image or video. This task can be used to identify key body locations, analyze posture, and categorize movements. It utilizes machine learning (ML) models that work with single images or video, and outputs body pose landmarks in image coordinates and in 3-dimensional world coordinates\cite{Google}\cite{Melo_2023}\cite{Google2}.

The Pose Landmarker task is available for various platforms, including Android, Python, and web-based applications, and provides code examples and guides for implementation. It offers configuration options such as setting the minimum confidence score for pose detection, the maximum number of poses that can be detected, and whether the task outputs a segmentation mask for the detected pose\cite{Google}\cite{Google2}.

The use of MediaPipe Pose has been explored in various research studies, including a paper that employed MediaPipe Pose to attain estimates of 2D human joint coordinates and an optimization method based on a humanoid model\cite{kim2023human}. Another study proposed a multi-scale pedestrian intent prediction method using 3D joint information as spatio-temporal representation, which does not require a detector and improves the run-time of previous approaches\cite{ahmed2023multi}.

Overall, MediaPipe Pose Landmarker provides a powerful and versatile framework for building perceptual pipelines and has been widely used in various applications, including human pose estimation, behavioral analysis, surveillance systems, and gesture recognition.

One such study by Camara et al. \cite{camara2018predicting} proposed a method for predicting pedestrian intention in a crosswalk scenario using a decision tree algorithm. The method used various features such as pedestrian speed, distance, and crossing time to predict the intention of the pedestrian. A related pose detection methodology by Radhakrishna et al. suggested a decision tree-based algorithm for pedestrian intention prediction in a real-time pedestrian detection system. The algorithm utilized features such as pedestrian location, speed, and angle of orientation as represented in figure 2 to predict the pedestrian's intent, albeit without further exploration \cite{radhakrishna2023economical}.

A study by Ge et al. proposed a pedestrian intention prediction system using a decision tree-based algorithm in a mixed traffic scenario. The system utilized various features such as pedestrian velocity, location, and direction to predict the pedestrian's intention, albeit in a nighttime setting \cite{ge2009real}. Jung et al. proposed optimizations to these systems by means of policy improvements \cite{jung2016improving}.

MonoPIC addresses several shortcomings present in current pedestrian intent detection systems, enhancing its effectiveness and applicability. Some notable shortcomings in existing systems that MonoPIC mitigates include:
\begin{itemize}
    \item Computational Efficiency: Many current systems struggle with computational efficiency, leading to delays in predicting pedestrian movement intent, especially in real-time scenarios. MonoPIC excels in computational efficiency, ensuring rapid and accurate predictions with minimal computational overhead. This improvement is crucial for applications requiring quick response times, such as autonomous vehicles and surveillance systems.

    \item Edge Deployment Challenges: Traditional systems may face challenges when deployed on resource-constrained IoT edge devices, impacting their feasibility for edge computing applications. MonoPIC is specifically designed for deployment on IoT edge devices, optimizing its resource utilization to operate seamlessly in edge computing environments. This addresses the limitations associated with deploying complex algorithms on devices with limited processing power and memory.

    \item Power Consumption Optimization: High power consumption is a significant concern for systems running on battery-powered devices or those aiming for sustainable operation. MonoPIC excels in power consumption optimization, ensuring efficient utilization of energy resources. This is crucial for extending the operational life of battery-powered devices and minimizing the environmental impact of continuous algorithmic processing.

    \item Speed and Responsiveness: In applications where real-time decision-making is critical, existing systems may exhibit latency issues, impacting their responsiveness. MonoPIC addresses this by demonstrating competitive speed and responsiveness, providing timely predictions of pedestrian movement intent. This is vital for applications such as autonomous navigation, where quick decision-making is imperative for safety and efficiency.
\end{itemize}

By addressing these shortcomings, MonoPIC can emerge as a promising solution that overcomes key challenges associated with current pedestrian intent detection systems, making it well-suited for deployment in diverse real-world scenarios.

In conclusion, decision trees have shown promise in pedestrian intent prediction for intelligent transport systems. The studies mentioned above demonstrate the effectiveness of decision tree-based algorithms in predicting pedestrian intent using various features. However, further research is required to develop more robust and accurate decision tree-based models for pedestrian intent prediction with emphasis on latency improvement.

\section{Concept Theory and Implementation Methodology}
In essence, the novel algorithm seeks to bridge the gap between custom IoT test beds designed for pedestrian obstacle avoidance and the low cost consumer market by implementing processing methodologies that can be run on devices with a low computational throughput as well as accomplish the same using a monocular 2-D camera. This will consequently allow us to improve the latency and the adoption of an IoT test-bed will introduce much needed limitations in energy consumption to more practically test the algorithm.

\subsection{Quaternion Transformations}

The proposed algorithm is designed to analyze a camera feed and detect humans in the image. It utilizes the Mediapipe \cite{lugaresi2019mediapipe} framework to isolate the skeletal structure of the human detected in the feed. The algorithm then collects three key pieces of information to predict the direction of movement of the human within the image frame.

First, it collects the angle of orientation of the human using the methods outlined by Radhakrishna et al. by extracting unit vectors denoting the position of each identified physical landmark, which allows us to derive our rotation matrix \cite{hamilton1866elements} using the camera look-at method \cite{prunier_2016} as

\begin{equation}\label{eq6}
\begin{split}
        R_{3,3} = \begin{bmatrix}
r_{00} &r_{01}  &r_{02} \\ 
r_{10} &r_{11}  &r_{12} \\ 
r_{20} &r_{21}  &r_{22}
\end{bmatrix} = \begin{bmatrix}
\widehat{x} &\widehat{y}  &\widehat{z} 
\end{bmatrix}
\end{split}
\end{equation}

where each column is a direct substitution of the constant values associated with the components of the unit vectors on the right hand side. The origin in these vector spaces is arbitrarily defined as the forward-bottom-left corner of the smallest cuboid encompassing the skeletal object, as visualized in figure 3. This arbitrary definition can be made as we are only looking for the rotation values rather than absolute depth.

\begin{figure*}
\centerline{\includegraphics[scale=0.4]{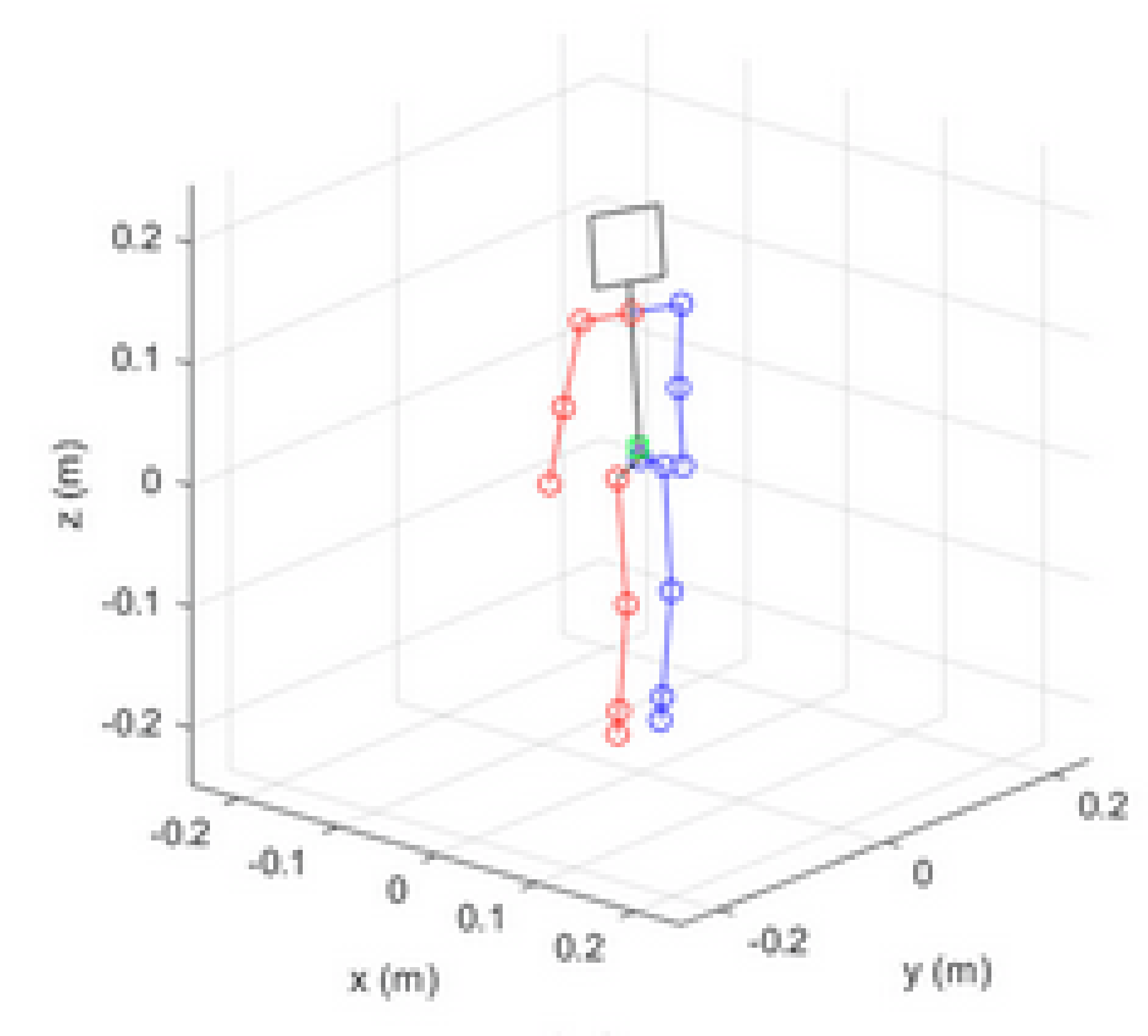}}
\caption{A representation of the manner in which the skeletal pose landmarks are visualized. Note the origins of the vector space\cite{kim2023human}}
\end{figure*}

Subsequently, the quaternion \textbf{Q} was obtained in a standard form from this rotation matrix \cite{hamilton1866elements} as 

\begin{equation}\label{eq7}
\begin{split}
        \textbf{Q} = a + bi + cj + dk
\end{split}
\end{equation}

where the coefficient values can be mapped according to equations 3 - 6 \cite{shuster1993survey}. 
\begin{equation}\label{eq8}
\begin{split}
        a = \frac{1}{2} \sqrt{|1 + r_{00} + r_{11} + r_{22}|} 
\end{split}
\end{equation}

\begin{equation}\label{eq9}
\begin{split}
        b = \frac{r_{21} - r_{12}}{4a}
\end{split}
\end{equation}

\begin{equation}\label{eq10}
\begin{split}
        c = \frac{r_{02} - r_{20}}{4a} 
\end{split}
\end{equation}

\begin{equation}\label{eq11}
\begin{split}
        d = \frac{r_{10} - r_{01}}{4a}
\end{split}
\end{equation}

Representing the quaternions as axis angles, 
\begin{equation}\label{eq12}
\begin{split}
        \textbf{Q} = cos\theta + sin\theta(xi + yj + zk)
\end{split}
\end{equation}

As such, rearranging this equation to our benefit, we can find $\theta$ to be
\begin{equation}\label{eq13}
\begin{split}
        \theta = \arccos (\frac{a}{\sqrt{a^{2} + b^{2} + c^{2} + d^{2}}})
\end{split}
\end{equation}

$\theta$ can then be transformed to obtain the angle of orientation in our frame of reference such that the  axis of reference was the horizontal across the camera feed window. Let's call this angle $\phi$.

\begin{equation}\label{eq14}
\begin{split}
        \phi = \theta \cdot \frac{180}{\pi} \cdot 4 - 180
\end{split}
\end{equation}

This angle provides a reference for the direction the human is facing. \\
The first constant of 180 by $\pi$ is multiplied to convert the quaternion angle from radians to degrees.
The second constant of 4 is and the subtraction of 180 were applied as linear error corrections after plotting the observed difference between $\theta$ and $\phi$ approximated to the nearest degree for 20 samples, and calculating the slope and constant, as emphasized by table 1 and figure 3. Evidently, the domain of values for this linear correction is [45, 90] with the range being [0, 180]. We can account for the slope in variation as quaternion axis angles are only equivalent to half their real world counterparts.
\\

\begin{table}[]
    \centering
    \begin{tabular}{c|c}
        \hline
        \text{$\theta$ (degrees)} & \text{$\phi$ (degrees)} \\
        \hline
        \text{88} & \text{175} \\
        \text{83} & \text{155} \\
        \text{78} & \text{135} \\
        \text{73} & \text{115} \\
        \text{68} & \text{95} \\
        \text{63} & \text{75} \\
        \text{58} & \text{55} \\
        \text{53} & \text{35} \\
        \text{48} & \text{15} \\
        \text{46} & \text{5} \\
        \hline
    \end{tabular}
    \caption{A sample of ten readings denoting the variation of phi by theta}
    \label{tab:my_label}
\end{table}

\begin{figure*}
\centerline{\includegraphics[scale=0.5]{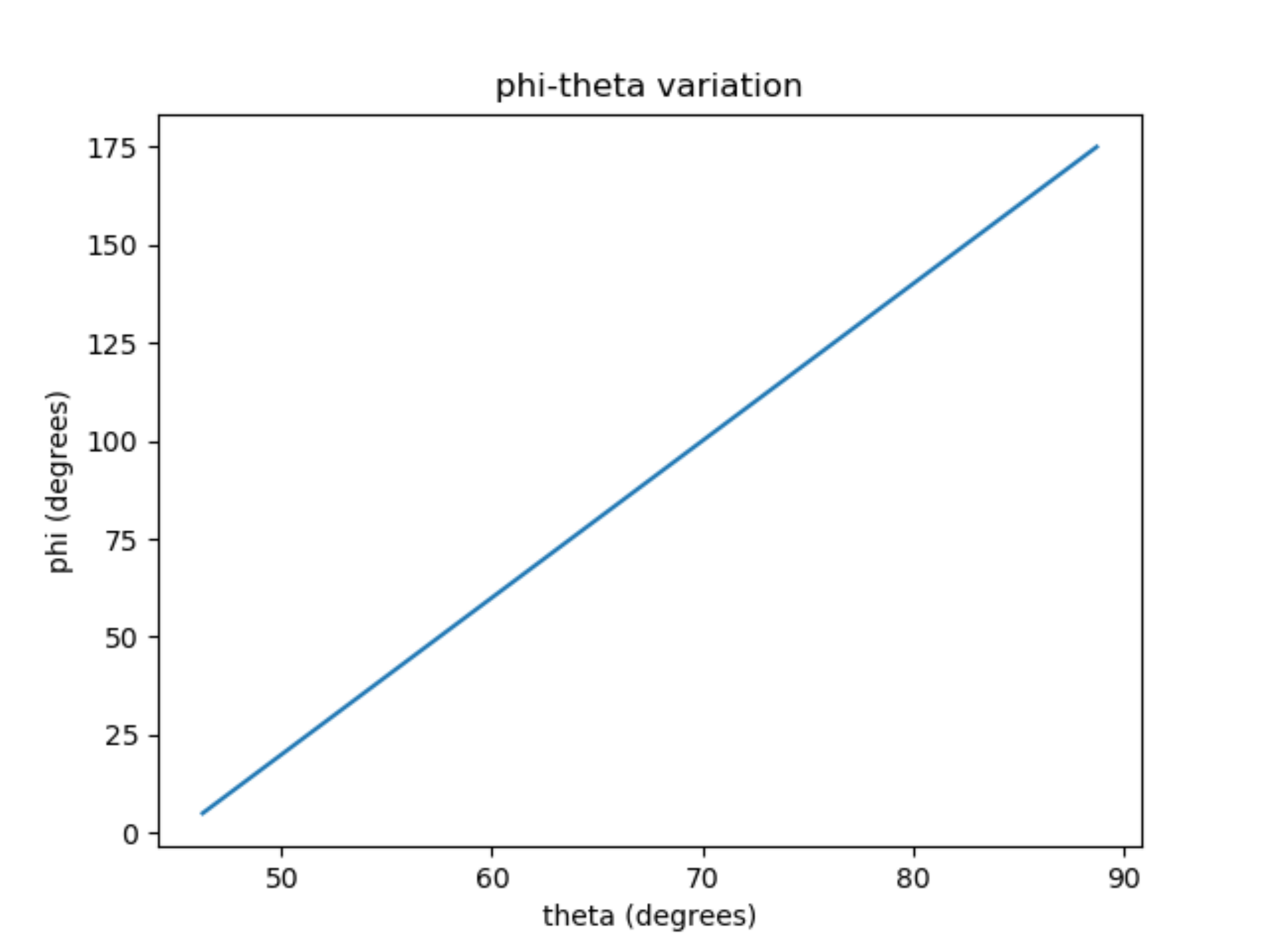}}
\caption{A plot showing the variation in readings, accurate to the degree with selective approximation up or down.}
\end{figure*}

Second, the algorithm measures the approach speed of the human in the X-axis of the image frame. This provides an indication of the horizontal speed of the human as they move within the image. Finally, the approach speed in the Y-axis of the image frame is measured. This provides an indication of vertical component of the velocity of the human on the frame. This calculation is made by dividing the delta in pixel locations by the framerate as shown in equation 10.

\begin{equation}\label{eq15}
\begin{split}
        \widehat{v} = \frac{\Delta [\widehat{x} \ \widehat{y} ]}{fps}
\end{split}
\end{equation}

where 'fps' denotes the frame rate per second. It's important to note the reference points for these position vectors with the underlying assumption being that the top-leftmost pixel on the frame acts as the origin.

Using this information, the algorithm can predict the direction of movement of the human within the frame. The predicted direction is based on the orientation angle of the human and the approach speeds along the X and Y axes.

Overall, the proposed algorithm represents a powerful tool for analyzing camera feeds and predicting the movement of humans within the image frame. With its use of the advanced Mediapipe framework and its collection of key pieces of data, the algorithm represents a significant step forward in computer vision and image analysis, although it is limited in its multi-human detection capability, unlike the approach employed by Prakash et al. \cite{prakash2014detecting}

To provide a birds-eye perspective on the novel algorithm, it functions by seeking out human objects in the frame, calculating a rotation matrix to determine said human's orientation in terms of his/her shoulder points by the camera look at method, calculating a quaternion from said rotation matrix and determining the angle being faced by the human object from the same with respect to the camera's horizon. This allows for a computationally lighter process to obtain the primary input for classification in the model. This reduces the abstractness and generality of deep learning algorithms employed for the same purposes as outlined previously. This is where the primary benefit of the proposed solution lies.

Furthermore, analysis of the human object being tracked on screen in terms of it's velocity on the 2D frame itself is performed by obtaining the perpendicular components of that velocity. This is done by analysing the x and y coordinates of specific points at multiple instances of time and calculating the derivative with respect to time at those two points. This process yields us enough information to model a decision tree that can help us predict the future location of the pedestrian in question within a timespan of a few seconds into the future. Intuitively, this solution should be able to help establish a quick control response to last minute pose changes in the scene by minimizing latency.

It is implied in this study that the human object being detected on frame will only by the one with the highest validation confidence. Given the fact, we've designed the test cases to involve only a single pedestrian at a time, although it is easily understandable how such an algorithm can be modified to detect multiple pedestrian objects by implementing multi-threading and distributing the computation among multiple cores.

 Stereoscopic depth estimation \cite{einecke2010two}, which involves analyzing the differences in the images captured by two cameras positioned a known distance apart, is the main method used to extract the landmarks in the images provided to the model. The depth information can then be used to calculate the 3D position and orientation of objects in the image. This approach is preferred in this case to attitude determination from vectors \cite{markley1988attitude} due to the flexibility of the system in case of an expansion into gait analysis for human-robot collaboration.

To calculate a quaternion from the data provided by the depth estimator and subsequently, the angle of orientation, we can make use of the camera look-ahead method, as demonstrated by Radhakrishna et al. \cite{radhakrishna2023economical}, expanded on from equation 1 through 10.

\subsection{ID3 Classification}

The ID3 algorithm \cite{hssina2014comparative} is a popular algorithm used for generating decision trees from labeled data. It was proposed by Ross Quinlan in 1986 and has since been widely used in various fields, including machine learning, data mining, and artificial intelligence \cite{quinlan1986induction}.

The ID3 algorithm recursively divides the dataset based on the features that provide the most information gain, resulting in a tree-like structure that represents a decision-making process. The resulting decision tree can be used for classification or prediction tasks.

The ID3 algorithm has several limitations, including a tendency to overfit on the training data and difficulty in handling continuous and missing data. However, it is a simple and effective algorithm that has paved the way for more sophisticated decision tree algorithms. Therefore, it was determined that this would be a good starting point to construct the decision tree.

Intuitively speaking, the movement of any object in the frame of a video can be classified into nine types, as highlighted by Radhakrishna et al., visualized in figure 5 - 

\begin{enumerate}
  \item Case 1 : Obliquely to the right, in the direction of the camera's motion.
  \item Case 2 : Obliquely to the left, in the direction of the camera's motion.
  \item Case 3 : Obliquely to the right, against the direction of the camera's motion.
  \item Case 4 : Obliquely to the left, against the direction of the camera's motion.
  \item Case 5 : Perpendicularly to the right, cutting across the path of the camera.
  \item Case 6 : Perpendicularly to the left, cutting across the path of the camera.
  \item Case 7 : Directly towards the camera.
  \item Case 8 : Facing away from the camera but still obstructing its path.
\end{enumerate}

\begin{figure*}
\centerline{\includegraphics[scale=0.5]{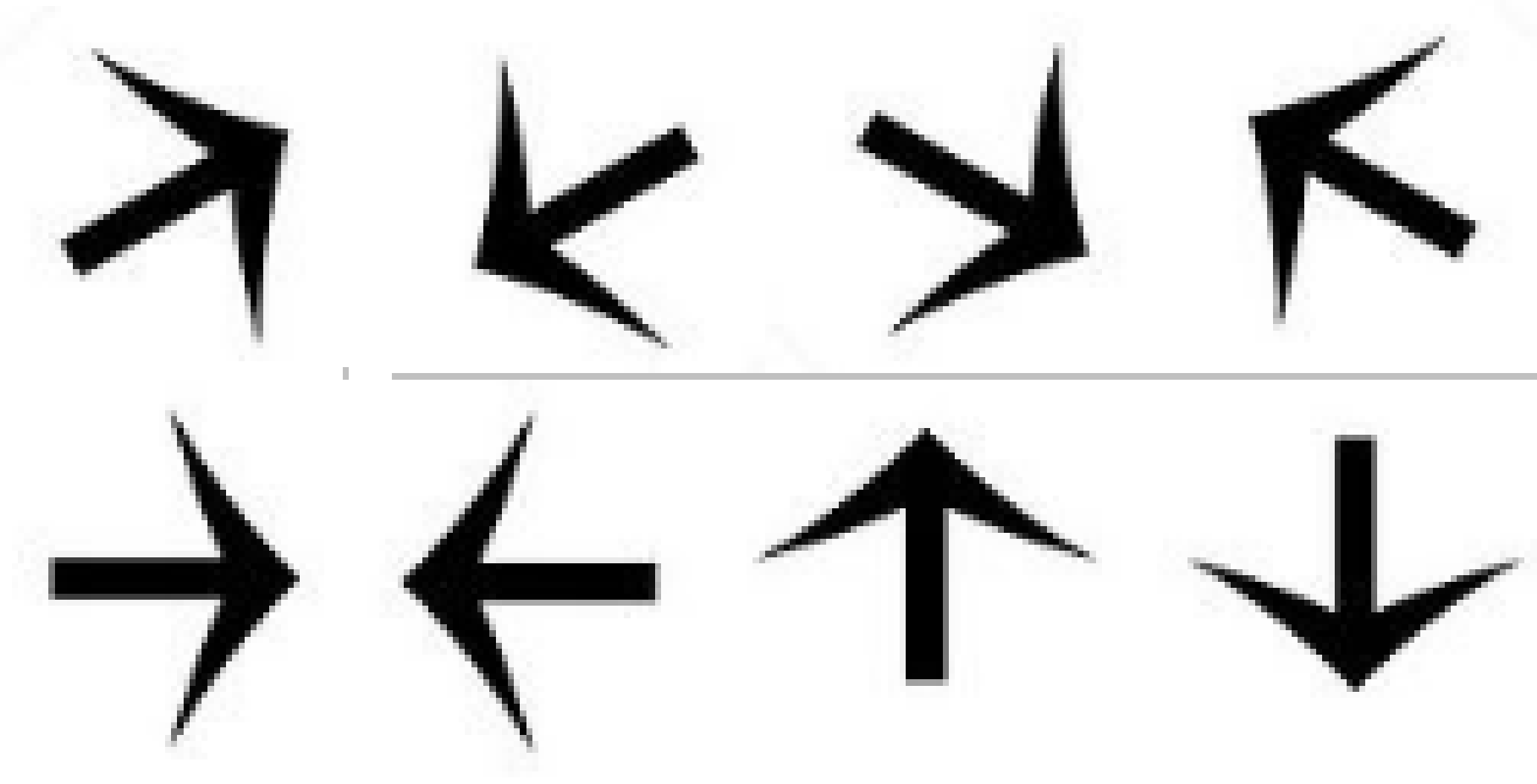}}
\caption{A representation of the directions of movement that can be observed on screen while evaluating a 2D frame, iterating from case 1 through 8}
\end{figure*}

The list of directions provided is comprehensive and covers all possible directions that an object could be moving on screen. These directions are based on the relative positions of the object and the edges of the image frame. Therefore, if an object is not moving or if its movement is not in any of these directions, then it would not be possible to describe its movement in terms of these directions.

It is important to note that the directions listed are relative to the camera's perspective and the orientation of the image frame. For example, if the camera is rotated 90 degrees, the directions of movement would also be rotated accordingly. Additionally, the directions listed assume that the object is moving in a straight line, and do not account for any complex movements such as curves or changes in velocity.

In many scenarios, pedestrians on roads tend to move relatively linearly, especially when walking along sidewalks or crossing streets.
Linear assumptions simplify the modeling and prediction processes, making it easier to implement and deploy in real-world applications. 

Pedestrians on roads often follow established paths, such as sidewalks or crosswalks, which may involve predominantly straight-line movements.
Sudden and complex movements are less common in typical pedestrian behavior on roads.

Most importantly, the scope of the detection itself is aimed at last minute obstacle avoidance, allowing us to estimate the path followed over distances in the order of magnitude of less than one meter. These paths are best approximated as straight lines.

In summary, the list of directions provided is exhaustive and covers all possible directions that an object could be moving on screen, based on the relative positions of the object and the edges of the image frame.

As per the algorithm to model the decision tree, it is necessary to calculate the information gain of each of the selected features, i.e.; the speed of the human on the frame in the horizontal direction, the same on the vertical and the angle of orientation. This can be performed as follows -

\begin{equation}
\text{Entropy}(S) = -\sum_{i=1}^{c} p_i \log_2 p_i
\end{equation}

where $S$ is the set of samples, $c$ is the number of classes, and $p_i$ is the proportion of samples in class $i$.

\begin{equation}
\text{Entropy}(S_A) = \sum_{v \in \text{Values}(A)} \frac{\left| S_{A=v} \right|}{\left| S \right|} \text{Entropy}(S_{A=v})
\end{equation}

where $A$ is the attribute being considered, $\text{Values}(A)$ is the set of possible values of attribute $A$, $S_{A=v}$ is the subset of samples in $S$ where attribute $A$ has value $v$, and $\left| S_{A=v} \right|$ is the number of samples in $S_{A=v}$.

\begin{equation}
\text{InformationGain}(A) = \text{Entropy}(S) - \text{Entropy}(S_A)
\end{equation}

where $\text{InformationGain}(A)$ is the information gain of attribute $A$ and $\text{Entropy}(S_A)$ is the entropy of the subset of samples in $S$ where attribute $A$ is considered.

Note that these equations can handle multi-class problems by using the general form of entropy in the same manner as described in equation 1.

Following this algorithm using the aforementioned features and classifying them into the nine rudimentary states indicating direction of movement on the frame, it's possible to obtain the decision tree as shown in figure 3, with the frame speed in the horizontal direction having the highest information gain of 0.022, the same in the vertical direction having an information gain of 0.014 and the angle of approach having an information gain of 0.004. These values were obtained by parsing the videos using the quaternion transformations mentioned earlier to obtain the frame speeds and orientations of the human objects.

\begin{figure*}[h!tbp]
\centering
\centerline{\includegraphics[scale=1, angle=90]{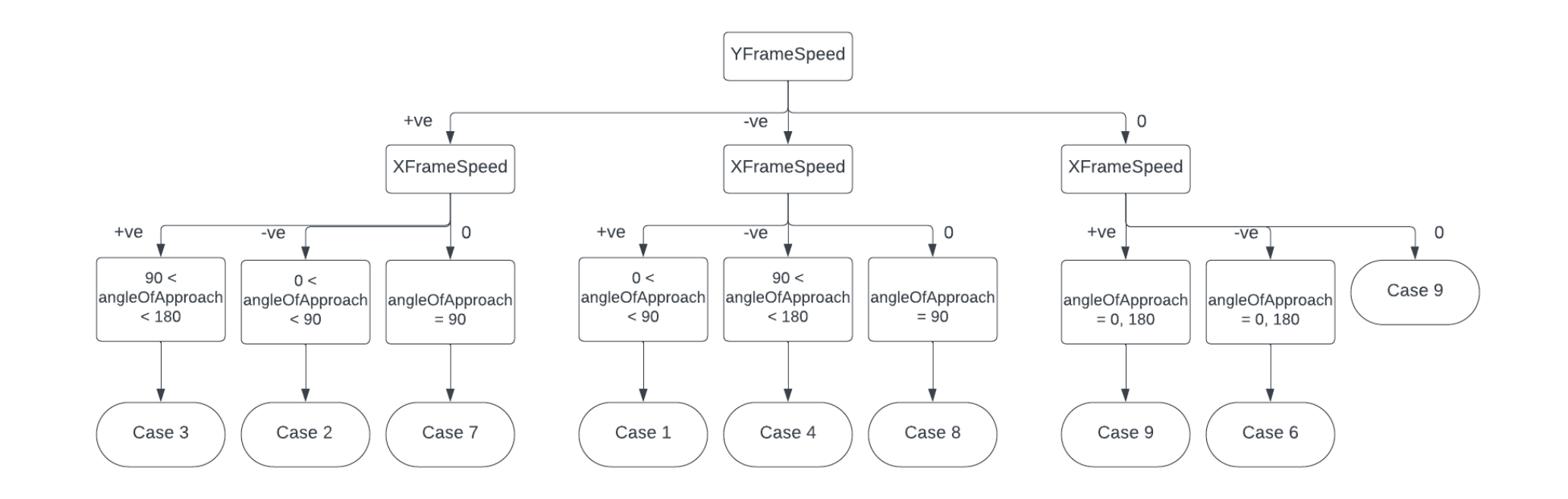}}
\caption{A representation of the decision tree obtained using the ID3 algorithm and the discretized dataset.}
\end{figure*}

\subsection{Algorithm}
The aggregated algorithmic flow for all variants of this novel concept follow the same general structure, as follows -

\begin{algorithm}
\caption{Calculate Q, $v_{l}$, $v_{r}$}\label{algo1}
\begin{algorithmic}[1]
\Require Video feed API $v$, time to future $t$
\Ensure Quaternion \textbf{Q}, average velocities of both shoulders $v_{l}$, $v_{r}$ 
\State $v$ = VideoCapture Class 
\State $prev\_pose$ = None
\State $prev\_time$ = None
\State // \textit{LOOP Process} :
\While{True}
        \State $img$ = $v$.frame // assigning an image requested from the API to the variable $img$ at the time of request
        \State $pose$ = poseDetector($img$) // detects the existence of a skeletal pose in the frame
        \State // getting the requested landmarks from the MediaPipe framework
        \If {$pose$ != None}
            \State $xl$, $yl$, $zl$ = $pose$.getLandmarks('Left Shoulder')
            \State $xr$, $yr$, $zr$ = $pose$.getLandmarks('Right Shoulder')
            \State $R$ = getRotationMatrix($xl$, $yl$, $zl$, $xr$, $yr$, $zr$) \\
            \State \textbf{Q} = calculateQuaternion($R$) // calculate quaternion from rotation matrix
            \State $time$ = $v$.getTimestamp() // get the timestamp of the current frame
            \If {$prev\_pose$ is not None}
                \State // calculate the average velocities of both shoulders
                \State $vl$ = calculateVelocity($prev\_pose$.getLandmarks('Left Shoulder'), $pose$.getLandmarks('Left Shoulder'), $prev\_time$, $time$, $t$)
                \State $vr$ = calculateVelocity($prev\_pose$.getLandmarks('Right Shoulder'), $pose$.getLandmarks('Right Shoulder'), $prev\_time$, $time$, $t$)
            \EndIf
        \State $prev\_pose$ = $pose$
        \State $prev\_time$ = $time$
        \EndIf 
\EndWhile
\end{algorithmic}
\end{algorithm}

In this version of the algorithm, two new variables have been added - $prev\_pose$ and $prev\_time$ - to keep track of the previous pose detected and the timestamp of the previous frame. A new input variable $t$ was also added, which represents the time in seconds between the current frame and the future frame to be used for calculating the velocities.
Additionally, two output variables were added - $v_{l}$ and $v_{r}$ - which represent the average velocities of the left and right shoulders over the specified time interval.
Finally, a new function was added called $calculateVelocity()$, which takes the landmarks of the shoulder in the current and previous frames, the timestamps of the current and previous frames, and the time interval $t$ as inputs and returns the average velocity of the shoulder over the time interval.
Note that the functions $getRotationMatrix()$, $calculateQuaternion()$, and $poseDetector()$ are assumed to be defined elsewhere in the code.

\subsection{Visualizations}

Three variants of this algorithm were developed to account for three use cases related to intelligent transport systems, with each successive one building on the last. Note the three parameters on the top-left of section of all the aforementioned demos. The first number represents the frame rate, the second one represents the latency in milliseconds and the third one represents the angle being detected by the algorithm with respect to the human's horizon.

\begin{itemize}
  \item The first variant predicts the path of movement one second ahead of time by classifying the pedestrian obstacle's future coordinates on the frame into 2D co-ordinates. After classification, the variant also checks for whether or not the future coordinates fall into the projection of a zone where the robot on which the camera is assumed to be mounted is projected to be an arbitrary time into the future. The prediction is one second ahead of time as this algorithm is intended to classify pedestrian obstacle movement behaviours short spans of time ahead of the present for quick reaction purposes. This variant also clearly states when a collision is imminent when the aforementioned conditions are met. \\
  
\begin{figure}[htbp]
\centering
\centerline{\includegraphics[scale=0.6]{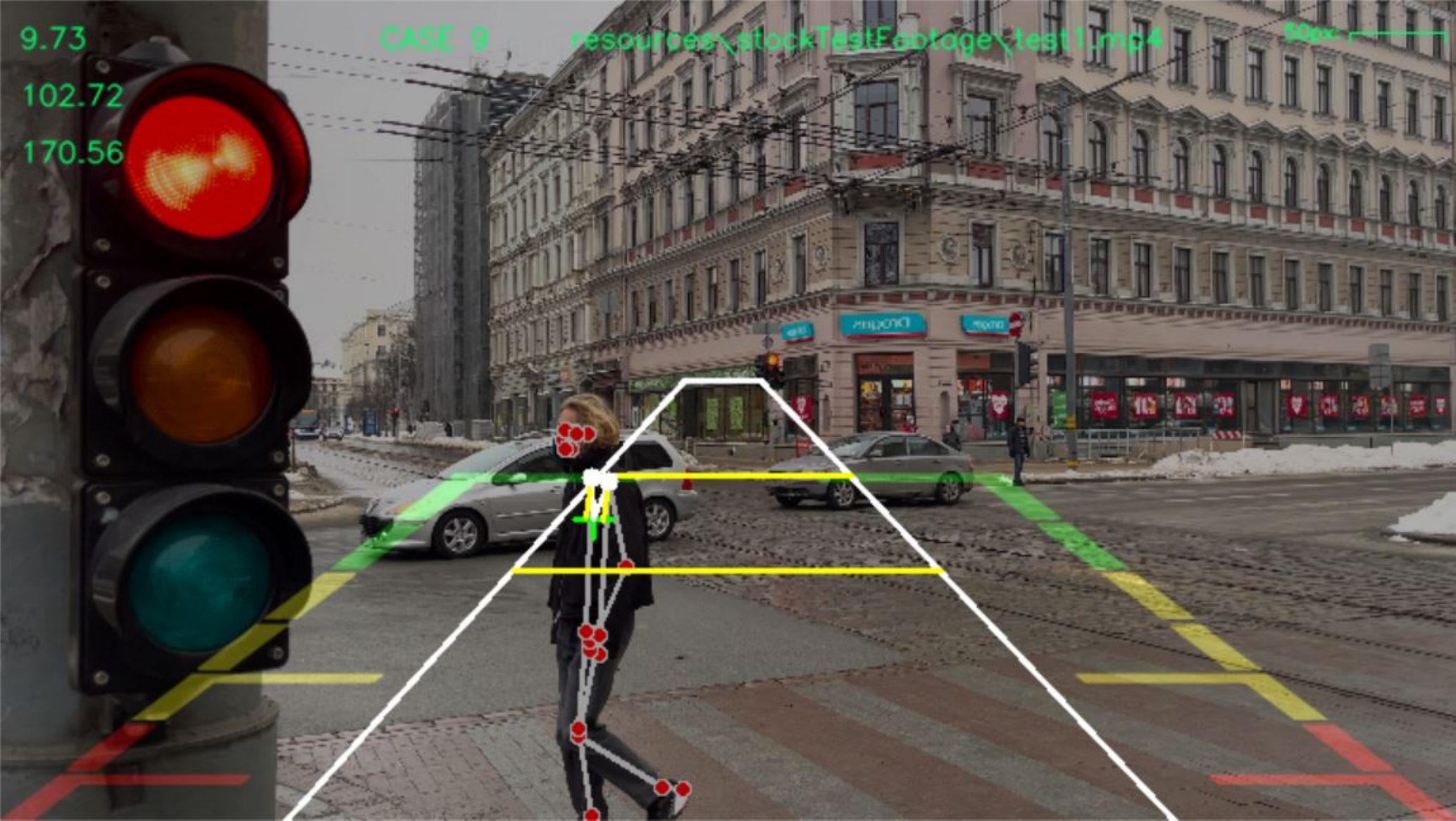}}
\caption{A demonstration of the first variant, capable of highlighting imminent collisions.}
 
\end{figure}
  
  \item The second variant extrapolates the predicted path in completion assuming no sudden changes in state for each frame based on data from the last defined one. This extrapolation is done over a frame containing a pre defined path overlay for the robot that the camera is assumed to be mounted on. This variant generates a full path from the location prediction source codes in the other two variants given in the directory and doesn't highlight the likelihood of collision. \\

\begin{figure}[htbp]
\centerline{\includegraphics[scale=0.6]{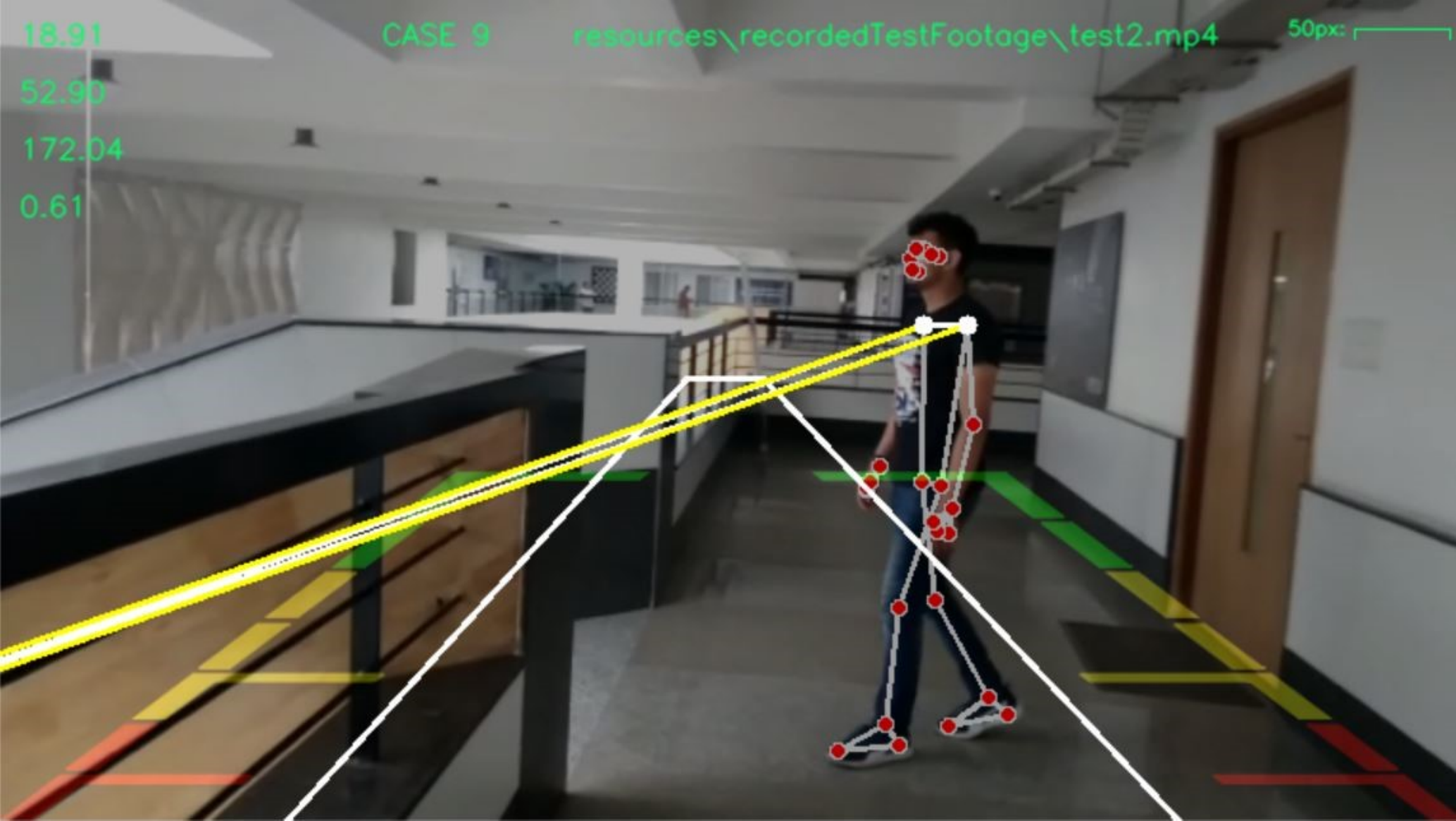}}
\caption{A demonstration of the second variant, capable of projecting a path overlay for the observed pedestrian. This representation highlights potential future path conflicts.}
\end{figure}
  
  \item The third variant classifies the predicted direction of movement of a pedestrian obstacle and visualizes it's projection on the frame. Data collection of various points mentioned in the code is conducted on this variant in order to visualize the range of accuracy of the model. This project intends to eliminate the usage of an additional ML pipeline to predict the same in order to reduce detection and classification latency as the use case targeted for this project is that of avoiding road accidents caused by autonomous robots due to failures in reacting on time to a sudden change in the movement intention of a pedestrian obstacle. As per preliminary literature surveys, modern standards for this detection speed is close to 200ms. This model is capable of performing the same in 50ms. Understandably, the numbers quoted in the previous point may appear to be outliers at first, due to which documentation will be provided defining the exact testing parameters. \\

\begin{figure}[htbp]
\centerline{\includegraphics[scale=0.6]{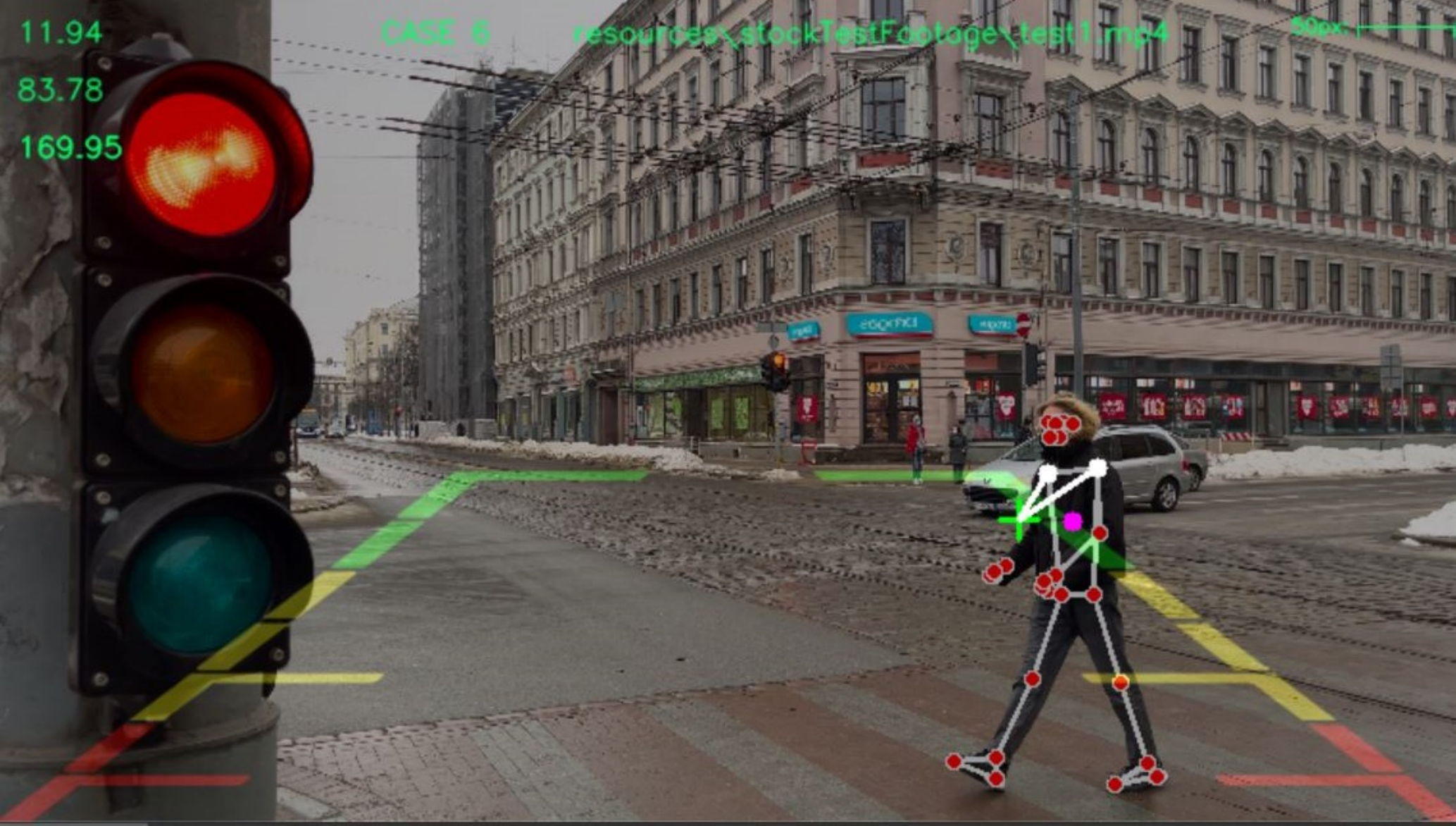}}
\caption{A demonstration of the third variant with a static overlay, ideal for mounting on an autonomous ground vehicle for last minute pedestrian avoidance.}
\end{figure}
  
\end{itemize}

\section{Results and Supporting Statistics}
Following is a brief about the conditions of testing, the datasets used and an overview of the results obtained.

\subsection{Dataset \& Testing Conditions}
The datasets constituted of human subjects (e.g.; the authors) walking on a flat surface in all nine of the manners specified in section 3, sub-section 2 as well as stock footage that satisfied those conditions. They consisted of seven videos of varying length and a collective count of 2328 frames recorded at an average of 15 frames per second. 

\subsection{Model Latency}

With respect to a recent work by by Mersch et al. and others \cite{mersch2021maneuver}\cite{wang2021human}, the authors proposed a temporal convolutional neural network that achieved high accuracy on a benchmark dataset and had a low latency estimated at the tens of milliseconds.

This score was acheived in a single stage in the pipeline. Applying the same principle to the novel algorithm presented in this paper, it needs to be noted that the BlazePose Convolutional Neural Network used in sequence with the decision tree here occupies 40 milliseconds of latency on the testbed employed (an average value determined after 20 iterations of testing). The average latency of the novel algorithm in the same span of testing was found to be 48 milliseconds, as observable in figure 6. Subtracting the two latencies gives us a decision tree latency of 8 milliseconds, as opposed to the temporal CNN approach's 10 plus milliseconds, with the result being achievable on a hardware implementation.

\begin{figure}[htbp]
\centerline{\includegraphics[scale=0.6]{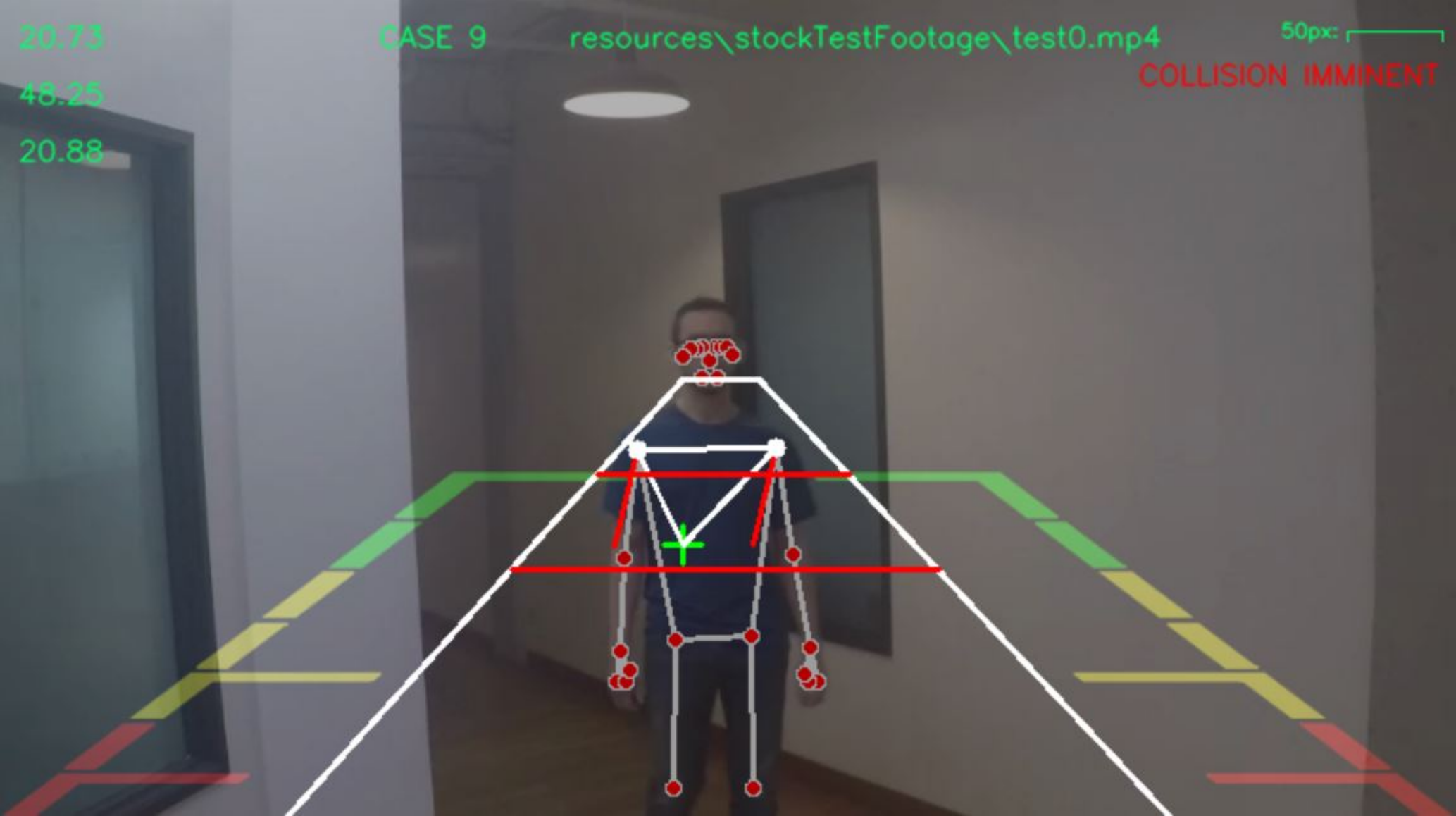}}
\caption{Note the three parameters on the top-left of section of this demo. The first number represents the frame rate, the second one represents the latency in milliseconds and the third one represents the angle being detected by the algorithm with respect to the human's horizon}
\end{figure}

\subsection{Model accuracy}

The decision tree's accuracy was found to diminish logarithmically with the reduction in error margin. Therefore, to evaluate the accuracy of the model, an arbitrary value of 50 pixels was taken to be the error margin. Considering that the length of the video frame was 432 pixels with the largest width of the captured cone being 10 meters, it can be extrapolated that this leaves us with a 1.15 meter distance between the actual location of the observed pedestrian on frame and their accurately predicted future location.

\begin{figure}[htbp]
\centerline{\includegraphics[scale=0.5]{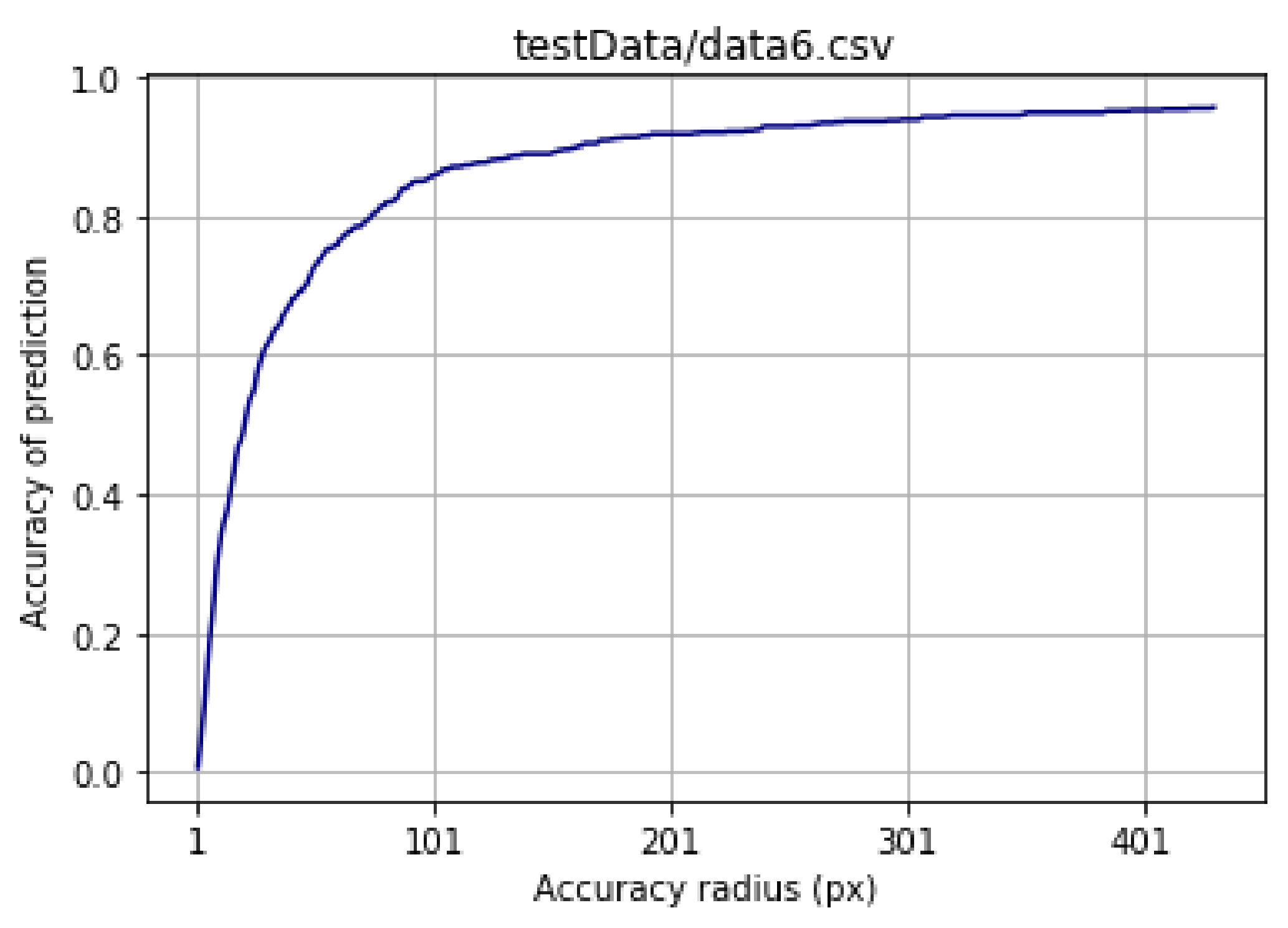}}
\caption{A plot of the accuracy of the model compared against the tolerable error margins prescribed to it on an arbitrary test case as a representative sample.}
\end{figure}

Assuming these conditions, the average prediction accuracy was found to be 83.56\%, as opposed to the 58.9\% achieved by Wang et.al. \cite{wang2021human}. While the two models were trained on different datasets, the custom dataset prepared to verify the novel algorithm in-house was captured at a very similar camera angle to the ETHZ dataset referenced by Wang et. al produced by Pellegrini et. al \cite{pellegrini2009you}, except with the key difference of capturing a single human at a time to conform to the capabilities of the Mediapipe BlazePose model's capacity to process a single subject at a time on a single thread. The publicly available stock footage used as part of this experiment was also verified to meet similar visual standards to ensure fairness in comparison. Nevertheless, this accuracy value was found post-deployment, thereby comparing the real world deployment accuracy of the models.

The novel algorithm also scored a variance of 0.0042 in its mean accuracy across seven files within the dataset (refer figure 9), indicating its reliability in deployment. 

\begin{figure}[htbp]
\centerline{\includegraphics[scale=0.5]{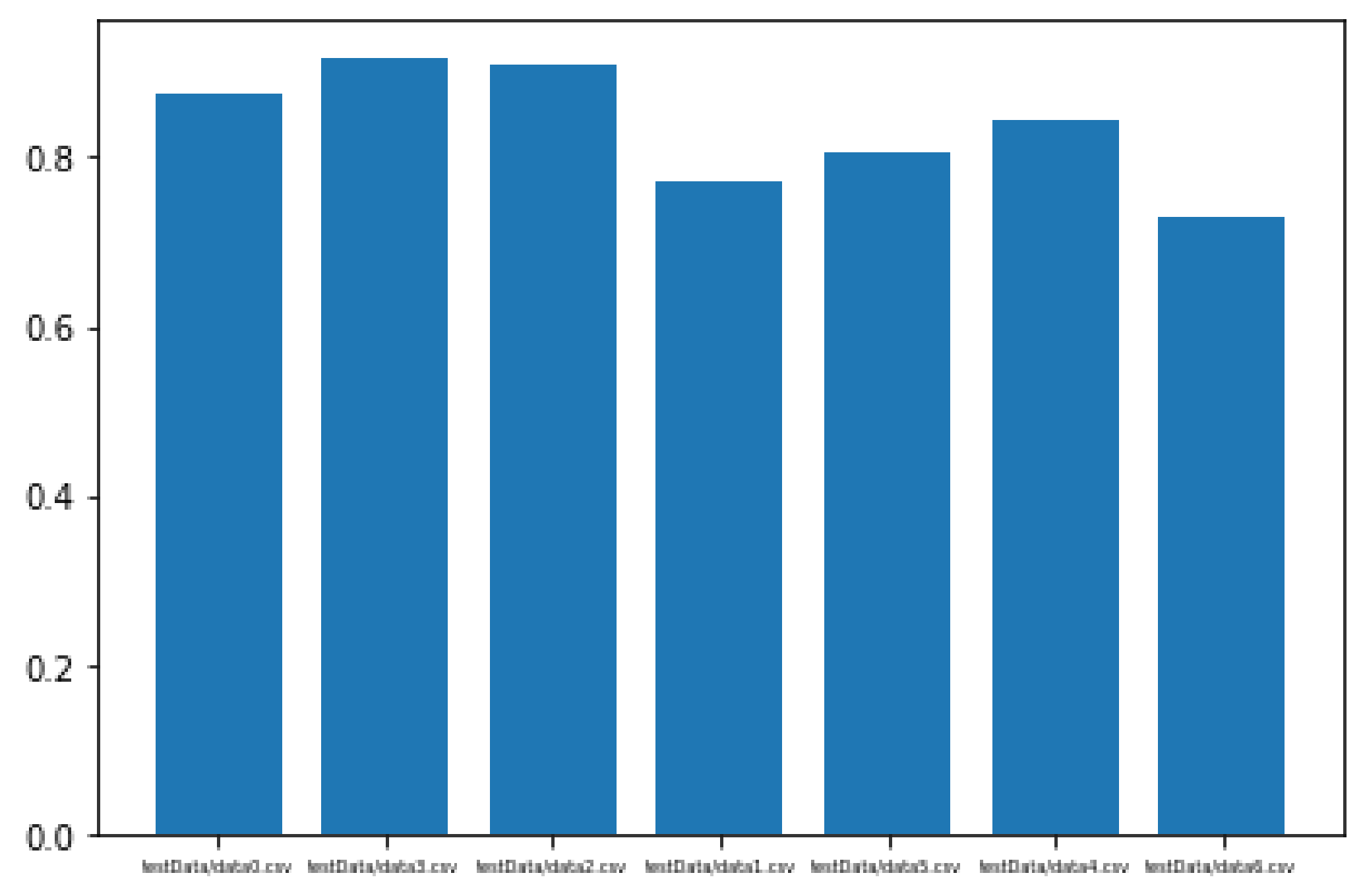}}
\caption{A plot of the mean accuracies of the model on all datasets within the sample space. Note the low variance implying the reliability of the model.}
\end{figure}

\subsection{System specifications}
This system was tested on a platform with the following specifications - 
\begin{itemize}
  \item Intel Core i5 7200U processor 
  \item 8GB RAM
  \item 512 GiB Solid State Drive
  \item External USB 2.0 30 FPS 2MP 'Passport' camera with a resolution of 1920x1080 and view angle of 110 degrees.
\end{itemize}

\subsection{Quantitative Analysis of Deployment}
In this section, we present a comprehensive quantitative analysis comparing our proposed framework utilizing quaternion-based decision trees against state-of-the-art deep learning-based monocular depth estimation algorithms, namely SQLDepth and DistDepth. Our evaluation focuses on two crucial performance metrics: the number of calculations required per prediction and the power consumption on the devices running these algorithms. We chose monocular depth maps as the benchmark for comparison due to their inherent suitability in real-world scenarios. 

\subsubsection{Choice of Monocular Depth Maps}
Monocular depth estimation techniques have gained prominence for their efficacy in extracting depth information from a single image, making them particularly relevant for real-time applications and resource-constrained devices. Unlike stereo or multi-view methods, monocular depth estimation requires only a single camera, thereby reducing hardware complexity. Additionally, monocular depth maps align with the lightweight philosophy of our proposed framework, ensuring compatibility with resource-constrained platforms.

\subsubsection{Experimental Setup}
We conducted our experiments on a diverse dataset that encompasses various pedestrian scenarios, including different lighting conditions, diverse backgrounds, and occlusions. The dataset was carefully curated to simulate real-world scenarios and ensure the generalizability of the proposed framework.

We implemented the framework using quaternion-based decision trees, leveraging the Python built-ins for tree construction and prediction. For comparison, we utilized two state-of-the-art monocular depth estimation algorithms, SQLDepth and DistDepth, implemented using popular deep learning frameworks. Calculations in this comparison accounted for additions, subtractions and multiplications carried out in one forward sweep on a frame of the video feed.

All of the involved modules were deployed on an edge of the following specifications -
\begin{itemize}
  \item AMD Ryzen 9 processor 
  \item 76 000 mWH battery
  \item 16GB RAM
  \item 1 TB Solid State Drive
\end{itemize}

\subsubsection{Calculations per Prediction}

Table 2 summarizes the number of calculations required per prediction for each algorithm. Our quaternion-based decision tree approach demonstrates a significant reduction in computational complexity compared to deep learning-based alternatives. This reduction is attributed to the efficiency of decision trees in handling feature spaces relevant to pedestrian intention classification.

\begin{table}[]
    \centering
    \begin{tabular}{c|c}
        \hline
        \text{Algorithm} & \text{Calculations per Prediction} \\
        \hline
        \text{MonoPIC} & \text{23} \\
        \text{SQLDepth} & \text{663 552} \\
        \text{DistDepth} & \text{1 000 000+} \\
        \hline
    \end{tabular}
    \caption{The number of calculations performed by equivalent depth sensing monocular algorithms without the MLP and softmax implied calculations compared to the entire calculation load of MonoPIC without pose detection}
    \label{tab:my_label}
\end{table}

The frame size taken for the calculation estimations of SQLDepth and DistDepth is 768 x 432.
It is important to note here that this solution is primarily applicable to cases where the reaction time for the robot or autonomous vehicle are massively reduced. In such a case, the curvature of the pedestrian obstacle's path can be reasonably assumed to be non-existent. This is why the proposed approach relies on linear extrapolations of the paths only.

\subsection{Power Consumption}
Power consumption is a critical consideration for real-time applications, particularly on devices with limited energy resources. In figure 13, we outline the estimated power consumption comparison for each algorithm after a runtime of 1 hour, revealing the efficiency of our proposed framework in minimizing energy requirements.

\begin{figure}[htbp]
\centerline{\includegraphics[scale=0.6]{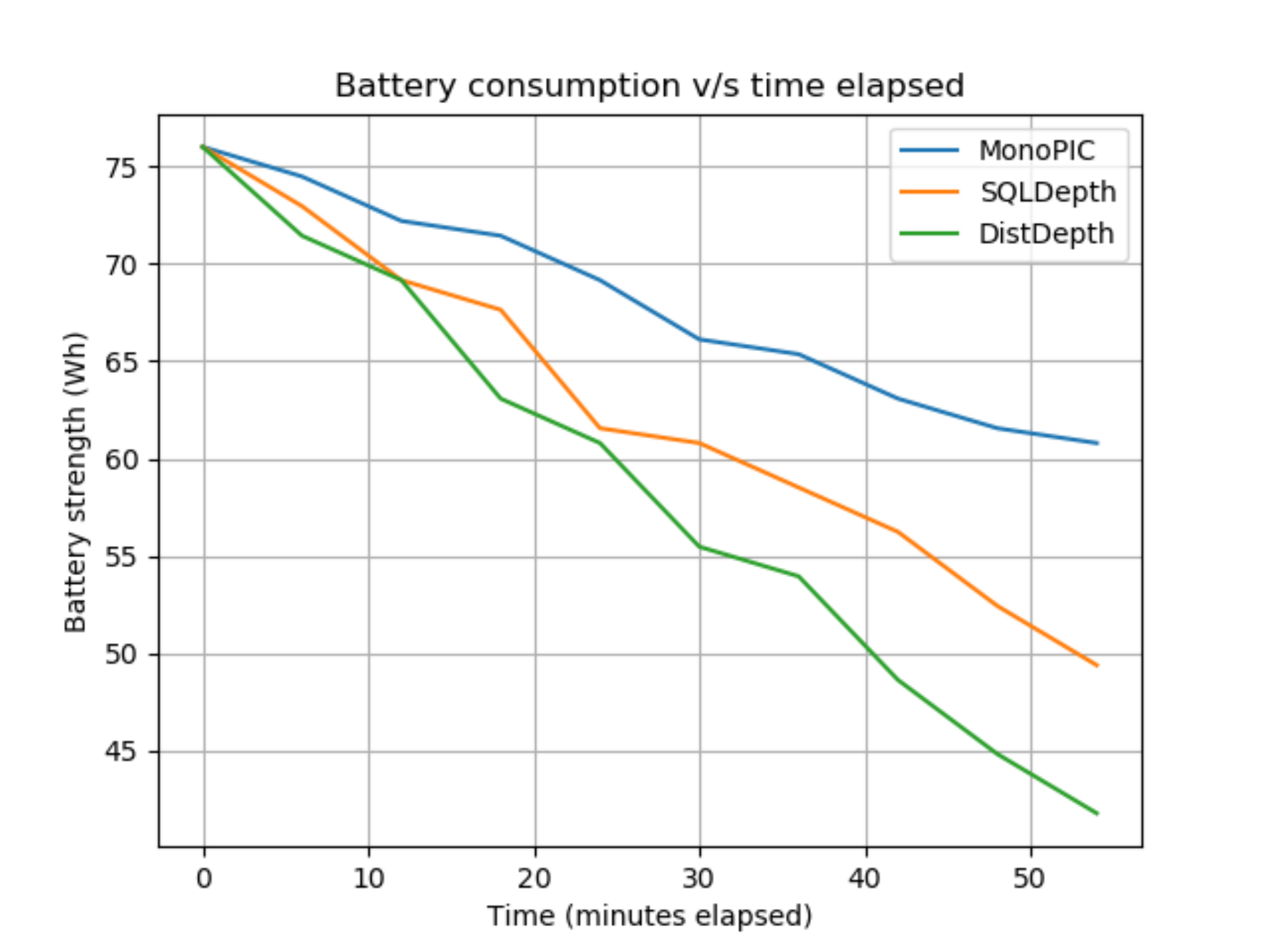}}
\caption{Power consumption readings of MonoPIC compared against SQLDepth and DistDepth for a duration of 1 hour with a refresh rate of 6 minutes.}
\end{figure}

The results in Tables 1 and 2 clearly demonstrate the superior efficiency of the proposed Lightweight, Monocular, Depth Agnostic Pedestrian Intention Classification Framework utilizing quaternion-based decision trees. The reduction in calculations per prediction and power consumption is particularly noteworthy, showcasing the practical viability of our approach for deployment on edge devices.

\subsubsection{Computational Efficiency and Power Consumption}
Quaternion decision trees exhibit remarkable computational efficiency, enabling real-time pedestrian intention classification with minimal computational overhead. This is in stark contrast to deep learning-based methods such as SQLDepth and DistDepth, which entail significantly higher computational demands.

The power consumption analysis further underscores the advantage of our proposed framework in resource-constrained environments. By leveraging quaternion-based decision trees, we strike a balance between accuracy and computational efficiency, resulting in a power-efficient solution for low-latency applications.

While our framework demonstrates promising results, it is essential to acknowledge certain limitations. The current evaluation focuses on a specific dataset, and generalization to diverse scenarios warrants further investigation. Additionally, ongoing research aims to explore enhancements to the decision tree model and the integration of adaptive learning mechanisms for improved performance.

In conclusion, our experimental evaluation establishes the efficacy of the proposed Lightweight, Monocular, Depth Agnostic Pedestrian Intention Classification Framework. By leveraging quaternion-based decision trees and comparing against state-of-the-art deep learning-based alternatives, we demonstrate superior computational efficiency and reduced power consumption. These findings position our framework as a compelling solution for real-time pedestrian intention classification on resource-constrained devices, paving the way for enhanced safety and responsiveness in various applications.

\subsubsection{Latency Comparison}
In the realm of latency, our proposed Lightweight, Monocular, Depth Agnostic Pedestrian Intention Classification Framework excels in providing swift and responsive predictions, outperforming deep learning-based counterparts. Quaternion decision trees showcase impressive efficiency, ensuring low-latency pedestrian intention classification. This stands in stark contrast to the latency profiles of SQLDepth and DistDepth, where the deeper architectures contribute to longer processing times, hindering real-time responsiveness.

\begin{table}[]
    \centering
    \begin{tabular}{c|c}
        \hline
        \text{Algorithm} & \text{Average Latency on test bench} \\
        \hline
        \text{MonoPIC} & \text{48.0 ms} \\
        \text{SQLDepth} & \text{4015.0 ms} \\
        \text{DistDepth} & \text{5472.0 ms} \\
        \hline
    \end{tabular}
    \caption{The average latency for frame return in MonoPIC far exceeds those of equivalent depth estimation models on the local test bench}
    \label{tab:my_label}
\end{table}

The latency analysis emphasizes the advantage of our framework in applications where timely decision-making is paramount. The efficient utilization of quaternion-based decision trees strikes a balance between prediction accuracy and low latency, making our solution well-suited for scenarios requiring quick and real-time insights into pedestrian movement intent.

However, it's crucial to acknowledge certain limitations in our evaluation. The latency comparison is based on a specific dataset, and broader applicability to diverse scenarios necessitates further exploration. Ongoing research endeavors aim to refine the decision tree model and explore adaptive learning mechanisms for continuous improvement in latency performance.

In summary, our experimental evaluation establishes the superiority of the proposed framework in achieving low-latency pedestrian intention classification. By leveraging quaternion-based decision trees and contrasting with deep learning-based alternatives, our solution demonstrates a significant reduction in processing time, positioning it as a compelling choice for applications demanding real-time responsiveness on resource-constrained devices.

\section{Conclusion}

In conclusion, the novel algorithmic implementation of a decision tree built on data evaluated from a dynamic human pose presents an interesting result in that it outperforms the closest comparable alternative in the task of pedestrian intent detection in terms of latency as well as accuracy. 
This can be attributed to the the lack of computational complexity in decision trees as opposed to the same in convolutional neural networks while the accuracy is ensured by limiting the training parameters by way of information gain evaluation using the ID3 algorithm. 
While deep learning based approaches work well with vision oriented tasks, they do tend to be computationally expensive to implement. At such times, approaches like this, involving the least number of parameters with the most impact can be highly practical, especially when their viable deployments to edge devices are in question.

\bibliography{sn-bibliography}% common bib file

\end{document}